\documentclass{article}

\usepackage{arxiv}
\usepackage[utf8]{inputenc} 
\usepackage[T1]{fontenc}    
\usepackage{hyperref}       
\usepackage{url}            
\usepackage{booktabs}       
\usepackage{amsfonts}       
\usepackage{nicefrac}       
\usepackage{microtype}      
\usepackage{lipsum}
\usepackage{graphicx}
\graphicspath{ {./images/} }

\usepackage{graphicx}
\usepackage{amsmath, amsfonts, dsfont}
\usepackage[bbgreekl]{mathbbol}
\usepackage{amsfonts}
\usepackage{enumitem}
\usepackage{footnote}
\usepackage{float}
\usepackage{hyperref}
\usepackage{pgfplots}
\pgfplotsset{compat=1.7}
\usepackage{times}
\usepackage{multicol}
\usepackage{multirow}
\usepackage{tikz}
\usepackage{lipsum} 
\usepackage{enumitem}
\setlist{nosep} 
\usepackage{color}
\usepackage{relsize}
\usetikzlibrary{calc}

\providecommand{\keywords}[1]
{
  \small	
  \textbf{\textit{Keywords---}} #1
}

\usepackage[ruled,vlined]{algorithm2e}

\usepackage{array}
\usepackage{booktabs}
\usepackage{makecell}
\usepackage{multirow}
\newcommand{\mc}{\multicolumn{1}{c}}

\usepackage{media9}
\usepackage{hyperref}

\usepackage{tikz}
\usepackage{forest}
\forestset{
  L1/.style={draw=black,},
  L2/.style={,edge={,line width=0.8pt}},
}

\usepackage{amsmath}
\usepackage{amssymb}
\usepackage{wasysym}

\newcommand{\circlesign}[1]{ 
    \mathbin{
        \mathchoice
        {\buildcirclesign{\displaystyle}{#1}}
        {\buildcirclesign{\textstyle}{#1}}
        {\buildcirclesign{\scriptstyle}{#1}}
        {\buildcirclesign{\scriptscriptstyle}{#1}}
    } 
}
\newcommand\buildcirclesign[2]{%
    \begin{tikzpicture}[baseline=(X.base), inner sep=0, outer sep=0]
    \node[draw,circle] (X)  {\ensuremath{#1 #2}};
    \end{tikzpicture}%
}

\usepackage{verbatim}
\usetikzlibrary{arrows,shapes}
\tikzstyle{format} = [draw, thin, fill=blue!20]
\tikzstyle{medium} = [ellipse, draw, thin, fill=green!20, minimum height=2.5em]
\usetikzlibrary{positioning,chains}

\title{Language Semantics Interpretation with an Interaction-based Recurrent Neural Networks}

\author{
 Shaw-Hwa Lo, Yiqiao Yin \\
  Statistics Department \\
  Columbia University \\
  \texttt{\{shl5, yy2502\}@columbia.edu} \\
}

\begin{document}
\maketitle
\begin{abstract}
Text classification is a fundamental language task in Natural Language Processing. A variety of sequential models is capable making good predictions yet there is lack of connection between language semantics and prediction results. This paper proposes a novel influence score (I-score), a greedy search algorithm called Backward Dropping Algorithm (BDA), and a novel feature engineering technique called the ``dagger technique''. First, the paper proposes a novel influence score (I-score) to detect and search for the important language semantics in text document that are useful for making good prediction in text classification tasks. Next, a greedy search algorithm called the Backward Dropping Algorithm is proposed to handle long-term dependencies in the dataset. Moreover, the paper proposes a novel engineering technique called the ``dagger technique'' that fully preserve the relationship between explanatory variable and response variable. The proposed techniques can be further generalized into any feed-forward Artificial Neural Networks (ANNs) and Convolutional Neural Networks (CNNs), and any neural network. A real-world application on the Internet Movie Database (IMDB) is used and the proposed methods are applied to improve prediction performance with an 81\% error reduction comparing with other popular peers if I-score and ``dagger technique'' are not implemented.
\end{abstract}


\section{Introduction}

\paragraph{Overview}{
    Artificial Neural Networks (ANNs) are created using many layers of fully connected units that are called artificial neurons. A ``shallow network'' refers to an ANN with one input layer while a ``deep network'' can have many hidden layers \cite{salehinejad2017recent}. In the architecture of a feed-forward ANN, each neuron has a linear component and a non-linear component that is defined using activation functions. Shallow networks provide simple architecture while deeper networks generate more abstract data representation \cite{salehinejad2017recent}. An important roadblock is the optimization difficulties caused by the non-linearity at each layer. Due to this nature, not many significant advances can be achieved before 2006 \cite{hinton2006fast, bengio2013advances}. Another important issue is is the generation of a big pool of datasets \cite{mahmood2021artificial, mahmood2021bridging}. A family of ANNs that have recurrent connections are called Recurrent Neural Networks (RNNs). These network architectures are designed to model sequential data for sequence recognition, classification, and prediction \cite{bengio1994learning}. There are a number of research in the literature about RNNs that are investigating discrete-time and continuous-time frameworks. In this paper, we focus on discrete-time RNNs. 
}

\paragraph{Problems in RNN}{
    The development of the back-propagation using gradient descent (GD) -- back-propagation through time (BPTT) -- has provided many great opportunities for training RNNs \cite{sutskever2011generating}. However, there are still a lot of challenges remain unsolved in modelling long-term dependencies \cite{salehinejad2017recent}. If we have a sequence of time-series features $X_t$, it is extremely challenging to detect the interaction relationship $X_t$ and $X_{t+c}$ have when $c$ is a large constant. This roadblock caused many inefficient training of BPTT which leads to extremely costly training procedure and lack of interpretation.
    
}

\paragraph{Problems in Text Classification Using RNN}{
    Recent years, there has been an exponential growth in the number of complex documents and texts that require a deeper understanding of prediction performance \cite{kowsari2019text}. A major drawback of using deep RNNs is the lack of interpretability for the prediction performance. Due to the nature of language, one word could appear in different forms (i.e. singular versus plural, present tense versus past tense) while the semantic meaning of each form is the same \cite{spirovski2018comparison}. Though many researchers have addressed some techniques to tackle the interpretation and semantic problems in natural language, few have been successful and most of the work still face limitations of feature selection \cite{spirovski2018comparison}. One famous technique is the N-gram technique (see Section \ref{subsec:language-modeling} for detailed overview). However, a major drawback of the N-gram technique is its difficulty at extracting features with long-term dependencies in the text document. 
}


\paragraph{Performance Diagnosis Test}{
    The performance of a diagnostic test in machine learning as well as Natural Language Processing in the case of a binary predictor can be evaluated using measures of sensitivity and specificity \cite{MANDREKAR20101315}. Due to the nature of activation functions (see Table \ref{tab:activationfct}) used in a deep learning architecture, we often times obtain predictors with continuous scale. This means we cannot directly measure the sensitivity, specificity, and accuracy rate which can cause inconsistency in the tuning process of the validating set and the robustness of the test set performance. In tackling this problem, a measure of using different range of cutoff points for the predictor to construct confusion table is proposed and this is called a Receiver Operating Characteristic (ROC) curve \cite{MANDREKAR20101315}. One major problem of using ROC curve to compute area-under-curve (AUC) values is that ROC treats sensitivity and specificity as equally important across all thresholds \cite{halligan2015disadvantages}. 
}

\begin{table}
    \centering
    \caption{\textbf{Famous Activation Functions.} This table presents three famous non-linear activation functions used in a neural network architecture. We use the ReLU as activation function in the hidden layers and the Sigmoid as activation function for the output unit. The activation functions are discussed in details in Apicella (2021) \cite{apicella2021survey} and we also compute the derivatives of these common activations in the table. }
    \begin{tabular}{llllll}
        \toprule
        \multicolumn{1}{l}{Name} & \multicolumn{1}{l}{Function} &  & \multicolumn{1}{l}{Figure} & & \multicolumn{1}{l}{Derivative} \\ 
        \hline
        Sigmoid & $\sigma(x)=\frac{1}{1+e^{-x}}$ &   &  
        \begin{tikzpicture}[baseline={(0,0.2)}]
         \draw (-1,0) -- (1,0);
         \draw (0,0) -- (0,1);
         \draw plot[domain=-1:1,variable=\x] ({\x},{1/(1+exp(-4*\x))});
        \end{tikzpicture} & &
        $\frac{\partial}{\partial x} \sigma(x) = \frac{e^{-x}}{(e^{-x} + 1)^2}$ \\
        \hline
        tanh & $\sigma(x)=\frac{e^x-e^{-x}}{e^x+e^{-x}} $ & 
        &  \begin{tikzpicture}[baseline={(0,0)}]
         \draw (-1,0) -- (1,0);
         \draw (0,-1) -- (0,1);
         \draw plot[domain=-1:1,variable=\x] ({\x},{tanh(4*\x)});
        \end{tikzpicture} & &
        $\frac{\partial}{\partial x} \sigma(x) = \frac{4e^{2x}}{(e^{2x} + 1)^2}$ \\
        \hline
        ReLU & $f(x) =\begin{cases}
        0 & ~\text{if}~ x<0 \\ 
        x & ~\text{if}~x \geq 0.
        \end{cases}$ &  & 
        \begin{tikzpicture}[baseline={(0,0.5)}]
         \draw (-1,0) -- (1,0);
         \draw (0,0) -- (0,1);
         \draw plot[domain=-1:1,variable=\x] ({\x},{ifthenelse(\x<0,0,\x)});
        \end{tikzpicture} & &
        $\frac{\partial}{\partial x} f(x) = \begin{cases}
        0 & ~\text{if}~ x<0 \\ 
        1 & ~\text{if}~ x>0 \\
        \text{not applicable} & ~\text{elsewhere}
        \end{cases}$ \\
        \bottomrule
    \end{tabular}
    \label{tab:activationfct}
\end{table}

\paragraph{Remark}{
    We start with discussion of the field of Natural Language Processing. In this field, we are facing data sets with sequential nature, which RNNs are designed to replace conventional neural networks architecture. A fundamental problem in this field is to study the language semantics using sequential data in text classification tasks. Many methods and models have been proposed, yet the conclusions are questionable. This is because in diagnostic tests of the prediction performance accuracy or AUC values are used as the benchmark methods to assess the robustness of a model. We show with both theoretical argument and simulation results that AUC exhibits major flaws in measuring how predictive a variable set or a model is at predicting target variable (or response variable). This is a fatal attack for AUC, because even with the correct features given AUC performs extremely poorly if the estimation of the true model is incorrect. We propose a novel I-score that increases while AUC value increases, but the proposed I-score does not subject to any attack from incorrect estimation of the true model. Accompanied with I-score, the ``dagger technique'' is proposed to further combine important features to form extremely powerful predictor. We regard this the major innovation for our paper. 
}

\paragraph{Contributions}{
    The proposed I-score exists in order to make good predictions. In supervised learning, suppose there are explanatory variables or features $X$ and target variables $Y$. The goal is to build a model (or a machine) to learn and estimate the true relationship between $X$ and $Y$. Dependent on different types of explanatory variables, many different approaches of neural networks are designed such as Artificial Neural Network (ANN), Convolutional Neural Network (CNN), and Recurrent Neural Network (RNN). It is a large consent to make the network deeper \cite{LeCun89, lecun2015deep, Krizhevsky2012, bengio2007scaling, Girshick2014, guyonetal2002, He2006, He2016, hinton2006fast, Huang2016} and rely on convolutional operation to extract features, yet the literature lacks exploration in truly understanding features by directly look at how features impact the prediction performance. If a machine $\hat{f}(\cdot)$ is trained, the prediction $\hat{Y}$ of the target variable $Y$ is a combination of $\hat{f}(\cdot)$ and $X$. It is not clear whether $\hat{Y}$ makes good prediction. Is it because of $\hat{f}(\cdot)$ or is it because of $X$? If it is because of $X$, how much impact does $X$ have on $Y$? The proposal of I-score is essentially to help us answer these questions cleanly without $\hat{f}(\cdot)$ to cloud our judgement. 
    
    I-score is derived from the theoretical prediction rate of explanatory features based on partition retention technique \cite{chernoffetal2009, lochernoffzhenglo2015, lochernoffzhenglo2016}. Based on I-score, Backward Dropping Algorithm (BDA) is also introduced to iteratively search for high-order interactions while omitting noisy and redundant variables (see detailed introduction in \ref{subsec:bda} Backward Dropping Algorithm). This paper extends the design of I-score and introduces a concept called the ``dagger technique'' (see discussion in \ref{subsec:interaction-based-features} Interaction-based Feature: Dagger Technique). I-score and BDA screens for important and predictive features directly while the ``dagger technique'' constructs a new feature based on the selected features that uses the local averages of the target variable $Y$ in training set. This powerful technique is able to efficiently screen for important features and constructs them into modules before they are fed into any neural network for training. 
    
    A theoretical novelty in this paper is the technical reasoning provided to show that I-score is a function of AUC. While AUC can only be used in the end of model, which means the impact of the features are couded by fitted model, I-score can be used anywhere in a neural network architecture. This paper shows proposed design of how I-score can be implemented in a type of neural network for text classification: Recurrent Neural Network (RNN). Though this paper focuses on RNN, similar design can be carried out implementing I-score with other types of neural networks.
    
    In practice, deep neural networks are generally considered ``black box'' techniques. This means that we typically feed in training data $X$ and $Y$, and predictions are generated by the network without clearly illustrating how $X$ affect the prediction. This inexplainability presents issues to end-users and prevents end-users to deploy a machine, sometimes well trained with high prediction performance, to live application. The entire production chain may hit a roadblock simply because end-users do not have sophisticated tools to understand the performance. I-score, in practice, shed lights to this problem. With direct influence measured by I-score, the impact that $X$ has on $Y$ can be computed without any incorrect assumption of the underlying model. The ``dagger technique'' can combine a subset of explanatory variables, which has two benefits. First, ``dagger technique'' features can be directly used to make predictions. Second, ``dagger technique'' features also presents end-users explainable and interpretable descriptions of the local average of target variable a partition has. In other words, if an observation in test set that an instance falls in a certain partition, we can directly read off the potential $Y$ value for this instance based on ``dagger technique''. This is practical novelty is not yet discovered in the literature.
}

\paragraph{Organization of Paper}{
    The rest of the paper is organized as follows. Section \ref{sec:a-novel-measure} starts with the presentation of a novel Influence Measure (i.e. Influence-score or I-score). This definition is introduced in paragraphs in \ref{subsec:iscore-confusion-auc} I-score, Confusion Table, and AUC. This definition, based on previous work \cite{chernoffetal2009, lochernoffzhenglo2015, lochernoffzhenglo2016}, is derived from the lower bound of the predictivity where we further discover the relationship between I-score and AUC. Next, we introduce an interaction-based feature engineering method called the ``dagger technique'' in paragraphs discussed in \ref{subsec:interaction-based-features} Interaction-based Feature: Dagger Technique. In addition, we present a greedy search algorithm called the Backward Dropping Algorithm (BDA) in paragraphs discussed in \ref{subsec:bda} Backward Dropping Algorithm, which is an extension of previous work \cite{chernoffetal2009, lochernoffzhenglo2015, lochernoffzhenglo2016}. We provide a toy example in paragraphs discussed in \ref{subsec:toy-example} Toy Example to demonstrate the application of I-score where we show with simulation that I-score is able to reflect true information of the features in ways that AUC cannot achieve. Section \ref{sec:application} Application discusses basic language modeling and the procedure of how to implement the proposed I-score and the ``dagger technique''. This section presents the basics of N-gram models and RNN in portions of paragraphs discussed in \ref{subsec:language-modeling} Language Modeling, an introduction of the dataset in paragraphs of \ref{subsec:imdb-data} IMDB dataset, and experimental results in paragraphs of \ref{subsec:result} Result.
}

\section{A Novel Influence Measure: I-score}
\label{sec:a-novel-measure}
This section introduces a novel statistical measure that assess the predictivity of the a variable set given the response variable (for definition of predictivity, see \cite{lochernoffzhenglo2015} and \cite{lochernoffzhenglo2016}). This I-score is formally introduced in the following.

Suppose the response variable $Y$ is binary (taking values 0 and 1) and all explanatory variables are discrete. Consider the partition $\mathcal{P}_k$ generated by a subset of $k$ explanatory variables $\{X_{b_1}, ..., X_{b_k}\}$. Assume all variables in this subset to be binary. Then we have $2^k$ partition elements; see the first paragraph of Section 3 in (Chernoff et al., 2009 \cite{chernoffetal2009}). Let $n_1(j)$ be the number of observations with $Y = 1$ in partition element $j$. Let $\bar{n}(j) = n_j \times \pi_1$ be the expected number of $Y = 1$ in element $j$. Under the null hypothesis the subset of explanatory variables has no association with $Y$, where $n_j$ is the total number of observations in element $j$ and $\pi_1$ is the proportion of $Y = 1$ observations in the sample. In Lo and Zheng (2002) \cite{lozheng2002}, the influence score is defined as 
\begin{equation}
    \label{eq:iscore-raw}
    I(X_{b_1}, ..., X_{b_k}) = \sum_{j \in \mathcal{P}_k} [n_1(j) - \bar{n}_1(j)]^2.
\end{equation}

The statistics I-score is the summation of squared deviations of frequency of $Y$ from what is expected under the null hypothesis. There are two properties associated with the statistics $I$. First, the measure $I$ is non-parametric which requires no need to specify a model for the joint effect of $\{X_{b_1}, ..., X_{b_k}\}$ on $Y$. This measure $I$ is created to describe the discrepancy between the conditional means of $Y$ on $\{X_{b_1}, ..., X_{b_k}\}$ disregard the form of conditional distribution. Secondly, under the null hypothesis that the subset has no influence on $Y$, the expectation of $I$ remains non-increasing when dropping variables from the subset. The second property makes $I$ fundamentally different from the Pearson's $\chi^2$ statistic whose expectation is dependent on the degrees of freedom and hence on the number of variables selected to define the partition. We can rewrite statistics $I$ in its general form when $Y$ is not necessarily discrete
\begin{equation}
    \label{eq:iscore}
    I = \sum_{j \in \mathcal{P}} n_j^2 (\bar{Y}_j - \bar{Y})^2,
\end{equation}
where $\bar{Y}_j$ is the average of $Y$-observations over the $j^\text{th}$ partition element (local average) and $\bar{Y}$ is the global average. Under the same null hypothesis, it is shown (Chernoff et al., 2009 \cite{chernoffetal2009}) that the normalized $I$, $I/n\sigma^2$ (where $\sigma^2$ is the variance of $Y$), is asymptotically distributed as a weighted sum of independent $\chi^2$ random variables of one degree of freedom each such that the total weight is less than one. It is precisely this property that serves the theoretical foundation for the following algorithm.

\subsection{I-score, Confusion Table, and AUC}
\label{subsec:iscore-confusion-auc}

Conventional practice in machine learning uses a variety of different performance metrics to examine the prediction procedure. Common metrics for measuring the robustness of a machine learning algorithms are defined below: \label{eq:metrics}
\begin{equation}
    \text{Accuracy}=\frac{\text{True positive} + \text{True negative}}{\text{True pos.} + \text{False pos.} + \text{False neg.} + \text{True neg.}}
\end{equation}
\begin{equation}
    \text{Sensitivity/Recall}=\frac{\text{True positive}}{\text{True positive} + \text{False negative}}
\end{equation}
\begin{equation}
    \text{Precision}=\frac{\text{True positive}}{\text{True positive} + \text{False positive}}
\end{equation}
\begin{equation}
    \text{Balanced Accuracy}=(\text{Sensitivity} + \text{Specificity})/2
\end{equation}
\begin{equation}
    \text{F1 Score} = \frac{2 \cdot \text{True positive}}{2 \cdot \text{True positive} + \text{False positive} + \text{False negative}}
\end{equation}
while the components of true positive, true negative, false positive, and false negative are presented in Table \ref{tab:conf_table}. 

As Table \ref{tab:conf_table} presented, these performance measure are well defined and suitable for two-class classification problem. However, the output from the forward pass of a neural network is generated by sigmoid activation. The sigmoid formula takes the form of $\sigma(x) = 1 / (1 + \exp(-x))$ and this definition bounds the output of a sigmoid function to be within the range of $[0,1]$. Additional non-linear activation functions can be found in Table \ref{tab:activationfct}, which are continuous prediction instead of binary class. This means there is not a unique confusion table. Due to this nature in neural network design, common practice also may choose ROC-AUC when measuring how accurate the predictor is. 

\begin{table}
    \centering
    \caption{\textbf{Confusion Table.} This table defines the components of a confusion table and the relationship between sensitivity/recall, precision, and F1 Score. For simplicity, denote true positive, false negative, false positive, and true negative to be $\alpha_1$, $\alpha_2$, $\alpha_3$, and $\alpha_4$, respectively. In generalized situation where predicted condition has more than two partitions, the cells of the confusion table can be simplified using the same $\alpha$'s directly. }
    \resizebox{0.8\textwidth}{!}{
    \begin{tabular}{ *{5}{|c} | }
      \cline{3-4}
      \mc{} & & \multicolumn{2}{c|}{Predicted condition} & \mc{} \\
      \cline{3-4}
      \mc{} & & \makecell{Condition \\ Positive } & 
        \makecell{Condition \\ Negative } & \mc{} \\
      \hline
      \multirow{5}{*}{\makecell{Actual \\ Condition}} & 
        \makecell{Positive } & 
        \makecell{True positive \\ $T_p = \alpha_1$ \\ \textit{(Correct)}} & 
        \makecell{False negative \\ $F_n = \alpha_2$ \\ \textit{(Incorrect)}} &
        \makecell{Sensitivity/Recall \\ Rate (RR) \\ $\frac{T_p}{T_p + F_n} \times 100\% = \frac{\alpha_1}{\alpha_1 + \alpha_2} \times 100\%$} \\
      \cline{2-5}
      & 
        \makecell{Negative \\ } & 
        \makecell{False positive \\ $F_p = \alpha_3$ \\ \textit{(Incorrect)}} &
        \makecell{True negative \\ $T_n = \alpha_4$ \\ \textit{(Correct)}} &
        \makecell{Specificity Rate \\ (SR) \\ $\frac{T_n}{T_n + F_p} \times 100\% = \frac{\alpha_4}{\alpha_3 + \alpha_4} \times 100\%$} \\
      \hline
      \mc{} & & \makecell{Precision/Positive \\ Predictive Value \\ (PPV) \\ $\frac{T_p}{T_p + F_p} \times 100\%$} &
        \makecell{Negative \\ Predictive Value \\ (NPV) \\ $\frac{T_n}{T_n + F_n} \times 100\%$} & \mc{} \\
      \cline{3-4}
      \mc{} & & \makecell{F1 Score \\ $\frac{2 \cdot T_p}{2 \cdot T_p + F_p + F_n} \times 100\%$} \\
      \cline{3-3}
    \end{tabular}
    }
    \label{tab:conf_table}
\end{table}

Based on the above confusion table, sensitivity can be computed according to the set of definitions in equation \ref{eq:metrics}. This means 
\begin{equation}
    \text{Sensitivity} = \frac{\alpha_1}{\alpha_1 + \alpha_2}
\end{equation}
\begin{equation}
    \text{Specificity} = \frac{\alpha_4}{\alpha_3 + \alpha_4}
\end{equation}
\begin{equation}
    1 - \text{Specificity} = 1 - \frac{\alpha_4}{\alpha_3 + \alpha_4} = \frac{\alpha_3}{\alpha_3 + \alpha_4}
\end{equation}
and the pair of statistics, (Sensitivity, 1 - Specificity), presents one dot on the figure which allows us to compute area-under-curve. 
\begin{figure}
    \centering
    \caption{\textbf{Mechanism between I-score Gain and AUC Gain.} This figure presents the mechanism of how I-score can increase AUC. There are four plots. The top left plot is a ROC curve with one particular pair of (1 - Specificity, Sensitivity). The top right plot presents sensitivity gain from I-score. The bottom left plot presents specificity gain from I-score. Both sensitivity and specificity are driving force of the AUC values because they move the dot up or left which then increase the area under curve (the blue area). The bottom right plot presents performance gain from both sensitivity and specificity. \textbf{In summary, implementation of using the proposed I-score can increase AUC by selecting the features raising both sensitivity (from part (i) of I-score, see equation \ref{eq:iscore-sens-spec-function}) and specificity (from part (ii) of I-score, see equation \ref{eq:iscore-sens-spec-function}).} }
    \includegraphics[width=\textwidth]{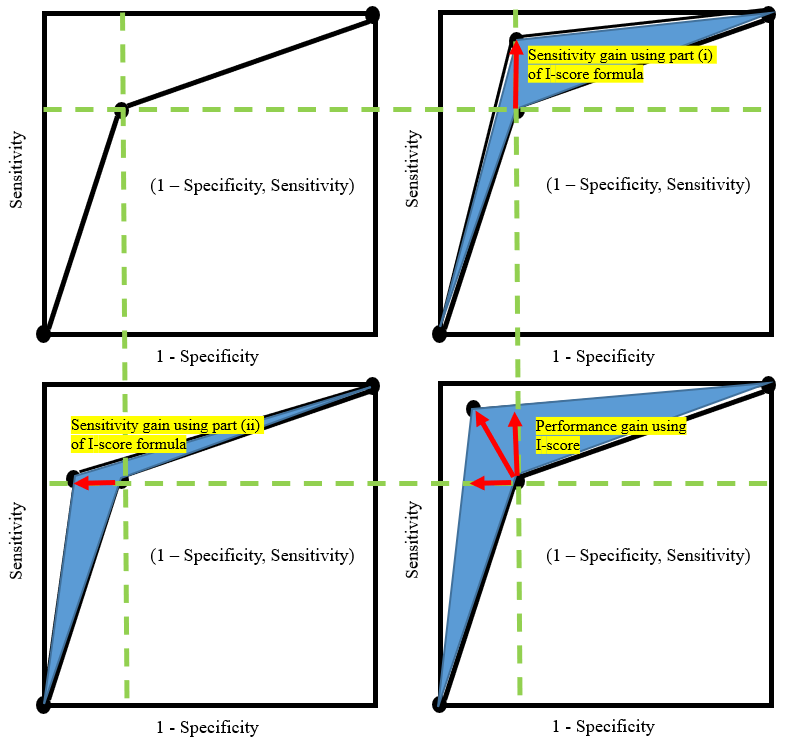}
    \label{fig:sample-auc}
\end{figure}

In order to compute I-score, partition is an important concept to understand. Since $X_1 \in \{0, 1\}$, the two partitions are simply $X_1 = 1$ and $X_1 = 0$. In the partition of $X_1 = 1$, $Y = 1$ with $\alpha_1$ observations and $Y = 0$ with $\alpha_3$ observations. In the partition of $X_1 = 0$, $Y = 1$ has $\alpha_2$ observations and $Y = 0$ has $\alpha_4$ observations. With these information, we can write the following
\begin{equation}
    \begin{array}{rcl}
        \bar{Y} &=& (\alpha_1+\alpha_2)/(\sum_{k=1}^4 \alpha_k) = (\alpha_1 + \alpha_2)/n, \\
        && \text{ if we let } n = \sum_{k=1}^4 \alpha_k \\
    \end{array}
\end{equation}
and next we can write out each terms in the proposed I-score formula (here each term is a partition). The first partition is  
\begin{equation}
    \begin{array}{rcl}
        \text{ when } j = 1, n_1^2(\bar{Y}_1 - \bar{Y})^2  
        &=& (\alpha_1 + \alpha_3)^2 (\frac{\alpha_1}{\alpha_1 + \alpha_3} - \frac{\alpha_1 + \alpha_2}{n})^2 \\
        &=& (\alpha_1 + \alpha_3)^2 \frac{[\alpha_1 n - (\alpha_1 + \alpha_2)(\alpha_1 + \alpha_3)]^2}{(\alpha_1 + \alpha_3)^2 n^2}, \text{ cancel } (\alpha_1 + \alpha_3)^2 \\
        &=& \big(\frac{\alpha_1 n - (\alpha_1 + \alpha_2)(\alpha_1 + \alpha_3)}{n}\big)^2, \\ 
        && \text{ dividing top and bottom (inside fraction) by } (\alpha_1 + \alpha_2) n \\
        &=& \big(\frac{\alpha_1/(\alpha_1 + \alpha_2) - (\alpha_1 + \alpha_3)/n }{1/(\alpha_1 + \alpha_2)}\big)^2, \text{ recall sensitivity } = \alpha_1/(\alpha_1 + \alpha_2) \\
        &=& \big(\frac{\text{Sensitivity} - (\alpha_1 + \alpha_3)/n }{1/(\alpha_1 + \alpha_2)}\big)^2, \text{ a function of sensitivity}
    \end{array}
\end{equation}
This means that if the variables with high sensitivity are selected, the I-score will increase which corresponds to an increase in AUC. Alternatively, the nature of I-score dictates to select highly predictive variables with high sensitivity which directly cause the ROC to move towards top left corner of the plot (see Figure \ref{fig:sample-auc}), i.e. resulting in higher AUC values.

Next, the second partition can be written as 
\begin{equation}
    \begin{array}{rcl}
        \text{ when } j = 2, n_2^2(\bar{Y}_2 - \bar{Y})^2 
        &=& (\alpha_2 + \alpha_4)^2 (\frac{\alpha_2}{\alpha_2 + \alpha_4} - \frac{\alpha_1 + \alpha_2}{n})^2 \\
        &=& (\alpha_2 + \alpha_4)^2 \frac{[\alpha_2 n - (\alpha_1 + \alpha_2)(\alpha_2 + \alpha_4)]^2}{(\alpha_2 + \alpha_4)^2 n^2}, \text{ cancel } (\alpha_2 + \alpha_4)^2 \\
        &=& \big(\frac{\alpha_2 n - (\alpha_1 + \alpha_2)(\alpha_2 + \alpha_4)}{n}\big)^2, \\ 
        &=& \big(\alpha_2 - (\alpha_1 + \alpha_2)(\alpha_2 + \alpha_4)/n\big)^2
    \end{array}
\end{equation}
Notice that under the scenario when the predictor $X_1$ is extremely informative, there can be very little observations for false negatives (which is $\alpha_2$). In other words, $\alpha_2 \approx 0$ when extremely predictive variable are present. In this case, the value of the second partition is completely determined by $(\alpha_1 + \alpha_2)(\alpha_2 + \alpha_4)/n$ which is largely dictated by the global average of the response variable $Y$ but scaled up with a factor of $(\alpha_2 + \alpha_4)$. A benefit from near zero $\alpha_2$ also implies that specificity can be high which is another direction to push AUC higher. 

As a summary, the two partitions that consists of the main body of I-score can be written as the following
\begin{equation}
    \label{eq:iscore-sens-spec-function}
    \begin{array}{rcl}
    \text{I}(Y, X_1) &=& \sum_{j=1}^2 n_j^2 (\bar{Y}_j - \bar{Y})^2 \\
        &=& n_1^2 (\bar{Y}_1 - \bar{Y})^2 + n_2^2(\bar{Y}_2 - \bar{Y})^2 \\
        &=& \underbrace{\bigg(\frac{\text{Sensitivity} - (\alpha_1 + \alpha_3)/n }{1/(\alpha_1 + \alpha_2)}\bigg)^2}_{\text{part (i), denote this as } \phi(\text{Sensitivity})} + \underbrace{\bigg(\alpha_2 - \frac{(\alpha_1 + \alpha_2)(\alpha_2 + \alpha_4)}{n}\bigg)^2}_\text{part (ii)}, \\
        && \text{ when } X_1 \text{ represents } Y \text{ completely}, \alpha_2 \approx 0 \\
        &\approx& \phi(\text{Sensitivity}) + \underbrace{\big(\bar{Y} \alpha_4\big)^2}_{\text{part (ii)}}, \text{ since } \bar{Y} = (\alpha_1+\alpha_2)/2, \text{because \text{Specificity}} = \frac{\alpha_4}{\alpha_3 + \alpha_4}, \\
        && \text{ so denote part (ii) as } \psi(\text{Specificity}) \\
    \Rightarrow \text{I}(Y, X_1)
    &\approx& \phi(\text{Sensitivity}) + \psi(\text{Specificity})
    \end{array}
\end{equation}
where it allows us to conclude:
\begin{itemize}
    \item \textbf{First, part (i) is a function of sensitivity.} More importantly, I-score serves as a lower bound of sensitivity. The proposed statistics I-score is high when sensitivity is high which means I-score can be used as a metric to select high sensitivity variables. A nice benefit from this phenomenon is that high sensitivity is the most important driving force to raise AUC values. This relationship is presented in the top right plot in Figure \ref{fig:sample-auc}.
    \item Second, part (ii) is a function of $\alpha_2$ which approximates to zero value when the variable is highly predictive. This leaves the second part to be largely determined by the global average of the response variable $Y$ but scaled up in proportion with the number of observations that fall in the second partition ($X_1 = 0$) which is the sum $\alpha_2 + \alpha_4$. An interesting benefit from this phenomenon is that the near zero $\alpha_2$ value jointly with part (i) implies that the specificity is high, which is another important driving force to raise AUC values. In other words, when the predictor has all the information to make good prediction performance, the value of $\alpha_2$ is expected to be approximately zero. In addition, the global mean of the true condition can be written as $\bar{Y} = \frac{1}{n}(\alpha_1+\alpha_2)$. Hence, this means that part (ii) can be rewritten as $(\bar{Y} \alpha_4)^2$ where $\alpha_4$ positively affect specificity, because specificity is $\alpha_4/(\alpha_3+\alpha_4)$. \textbf{Thus, part (ii) is a function of specificity.}
    \item Third, I-score is capable of measuring variable set as a whole without making any assumption of the underlying model. However, AUC is defined between a response variable $Y$ and a predictor $\hat{Y}$. If a variable set has more than one variable, some underlying assumption of the model need to be made -- we would need $\hat{Y} := f(X_1, X_2, ..)$ -- in order to compute AUC value.
\end{itemize}

\paragraph{Remark}{
    In generalized situation, the confusion table in Table \ref{tab:conf_table} may have more than two partitions for predicted condition. In any scenarios that the table has three or more partitions, they can be written into two partitions using a threshold. For example, instead of positive or negative, there can be predicted condition to take values of $\{1, 2, ..., K\}$. In this case, any number can be taken excluding $1$ and $K$ as threshold to reduce $K$ levels back into $2$ levels. Suppose ``2'' is used as a threshold, the partitions greater than $2$ can be one partition and the partitions less than or equal to 2 can be one partition. This allows to reduce $K$ levels into $2$ levels. The same proof in equation \ref{eq:iscore-sens-spec-function} would follow.
}

\subsection{An Interaction-based Feature: Dagger Technique}
\label{subsec:interaction-based-features}

The concept of interaction-based Feature is initially proposed in Lo and Yin (2021) \cite{lo2021interaction}. In their work, the authors defined an interaction-based feature that is used to replace the construction of using filters in designing Convolutional Neural Networks (CNNs). The conventional practice relies on pre-defined filters and these filters are small 2-by-2 or 3-by-3 window that are designed to capture certain information based on prior knowledge. The art of using interaction-based feature to create novel features within a 2-by-2 or 3-by-3 window in an image is to allow the data rather than meaningless filters to indicate the predictive information in the image. These new features are denoted as $X^\dagger$'s, and hence the name ``dagger technique''. The rest of this subsection formally define this method of using partitions to define novel features.

A major benefit for using the proposed I-score is the partition retention technique. This is a feature engineering technique that helps us to preserve the information of a variable set and convert it into one feature. Since ROC AUC cannot be directly computed between a response variable and the potential variable set, common procedure tends to fit a model first before AUC is computed. This is a very costly method for the following two reasons. First, the fitting of a regression or a classification model can be very costly to train. Second, the model fitting procedure cannot guarantee the prediction results of the final predictor. If the AUC value is low, there is no solution to distinguish whether the poor AUC result comes from model fitting or variable selection. It is shown with evidence in simulation that there can be extremely poor AUC values when the correct features are present with an incorrect estimation of the model (see subsection \ref{subsec:toy-example} for simulation discussion and Table \ref{tab:sim-result-one-module} for simulation results). We consider this the major drawback of using ROC AUC. 

To tackle this problem, a proposed technique is to use partition retention. These new features are denoted as $X^\dagger$'s and hence this method is named the ``dagger technique''. Now we introduce this technique as follows. Suppose there is a supervised learning problem and there are explanatory variables $X$ and response variable $Y$. Suppose $X$ has partitions size $k$. A novel non-parametric feature can be created using the following formula
\begin{equation}
    \label{eq:interaction-based-features}
    X^\dagger := \bar{Y}_j, \text{ while } j \in \{1, 2, ..., k\}
\end{equation}
where $k$ is the size of the total partitions formed by $X$. For example, suppose there is $X_1 \in \{1,0\}$ and $X_2 \in \{1,0\}$. Then the variable set $\{X_1, X_2\}$ has 4 partitions, i.e. computed using $2^2 = 4$. In this case, the running index for notating the partition $j$ can take values $\{1,2,3,4\}$. Then, based on this variable set $\{X_1, X_2\}$, a new feature can be created called $X^\dagger_{X_1,X_2}$ that is a combination of $X_1$ and $X_2$ using partition retention. Hence, this new feature can be defined as $X^\dagger_{X_1,X_2} := \bar{Y}_j$ while $j \in \{1,2,3,4\}$ as discussed above. The results of this example is summarized in tabular form (see Table \ref{tab:dagger-for-small-example}).

\begin{table}
    \centering
    \caption{\textbf{Interaction-based Engineer: ``Dagger Technique''.} This table summarizes the construction procedure of $X^\dagger$ (the ``dagger technique''). Suppose there is a variable set $\{X_1, X_2\}$ and each of them can take values in $\{0, 1\}$. The $X^\dagger$ can be constructed and the values of this new feature is defined using the local average of the target variable $Y$ based on the partition retained from the variable set $\{X_1, X_2\}$. Here the variable set $\{X_1,X_2\}$ produces 4 partitions. Hence, the $X^\dagger$ can be defined according to the following table. In test set, the target variable (or response variable) $Y$ is not observed, so the training set values are used. Hence, the reminder is that in generating test set $X^\dagger$ we use $\hat{y}_j$'s from training set. }
    \[
    \begin{array}{lcc}
        \toprule
        \text{Training set}: \\
        X^\dagger & X_1 & X_2 \\
        \hline
        \bar{y}_1 = \mathbb{E}(Y|X_1=1,X_2=1) & 1 & 1 \\
        \bar{y}_2 = \mathbb{E}(Y|X_1=1,X_2=0) & 1 & 0 \\
        \bar{y}_3 = \mathbb{E}(Y|X_1=0,X_2=1) & 0 & 1 \\
        \bar{y}_4 = \mathbb{E}(Y|X_1=0,X_2=0) & 0 & 0 \\
        \bottomrule
    \end{array}
    \rightarrow
    \begin{array}{lcc}
        \toprule
        \text{Test set}: \\
        X^\dagger & X_1 & X_2 \\
        \hline
        \bar{y}_1 \text{ (generated from training set}) & 1 & 1 \\
        \bar{y}_2 \text{ (generated from training set}) & 1 & 0 \\
        \bar{y}_3 \text{ (generated from training set}) & 0 & 1 \\
        \bar{y}_4 \text{ (generated from training set}) & 0 & 0 \\
        \bottomrule
    \end{array}
    \]
    \label{tab:dagger-for-small-example}
\end{table}

\subsection{Discretization}
\label{subsec:discretization}

The partition retention technique requires the partition space of the subset of variables in I-score formulation to be countable. In other words, the partition is necessary to form and it is can be defined by the subset $\tilde{X}_b$ disregard how many variables are selected in this group $b$. If each variable takes values $\{0, 1\}$, then we have $2^k$ partitions for this subset $\tilde{X}_b$. This is, however, not always guaranteed from practice. In some situation, there can be variables with many unique values that can be considered continuous. To avoid sparsity problem\footnote{Sparsity problem arises when there are too many unique levels with lack of training sample size in the data. Without proper discretization or grouping technique, we would have a partition with zero observations in it which can slow down the computation of the statistics I-score.}, we would need to discretize the continuous variable into discrete variable. The algorithm is presented in Algorithm \ref{alg:discretization}.

Suppose any explanatory variable $X_j$ has $l$ unique levels while $j$ in $1, 2, ..., p$. That means, we can use order statistics to write out all the unique levels using $X_{j,(1)}, ..., X_{j,(l)}$. To discrete $X_j$ from $l$ levels into two levels, we need to choose a cutoff to compare all levels against this threshold (call it $t$), i.e. binary output based on $t$ would be $\mathds{1}(X_j > t)$, in order to create a new variable that takes only values in $\{0,1\}$. This threshold $t$ can take any value in the unique levels, i.e. $\forall t \in \{X_{j,(1)}, ..., X_{j,(l)}\}$ and at each $t$ we can compute I-score using equation \ref{eq:iscore}. The best threshold $t^*$ would be the candidate that maximizes I-score. The objective function can be stated as the following:
\begin{equation}\label{eq:discretization}
    t^* = \{t: \max_t \text{I}(Y, X_{j,t})\} = \{t: \max_t \frac{1}{n\sigma^2} \sum_{s_t=1}^2 n_{s_t}^2 (\bar{Y}_{s_t} - \bar{Y})^2 \}
\end{equation}
while $s_t$ is dependent on the threshold $t$ and can take $\{1, 2\}$, because the partition is constructed based on an indicator function, i.e. $\mathds{1}(X_j > t)$ and can only be two partitions.

\begin{algorithm}[H]
\caption{\textbf{Discretization.} Procedure of Discretization for an Explanatory Variable \label{alg:discretization}}
\emph{Define unique levels:} $X_{j,(1)}, X_{j,(2)}, ..., X_{j,(l)}$ \\
\emph{Initialize:} set $t^* = 0$, and $\text{I}^* = 0$ \\
for $t$ in unique levels:

\For{$t$ in unique levels:}{
    Compute $\text{I}_t = \frac{1}{n\sigma^2} \sum_{s_t=1}^2 n_{s_t}^2 (\bar{Y}_{s_t} - \bar{Y})^2$ \\
    \If{$\text{I}_t > \text{I}^*$}{
        set $t^* = t$
    } \Else{
        continue
    }
}
\emph{Conversion:} Use indicator function to convert $X_j$ into binary form according to threshold $t^*$, i.e. $\mathds{1}(X_j > t^*)$.
\end{algorithm}

\subsection{Backward Dropping Algorithm}
\label{subsec:bda}

In many situation, a variable set has noisy information that can damage the prediction performance. In this case, we recommend Backward Dropping Algorithm to omit the noisy variables first before we do machine learning or using the ``dagger technique'' which refers to the feature engineering technique in equation \ref{eq:interaction-based-features}.

The Backward Dropping Algorithm is a greedy algorithm to search for the optimal subsets of variables that maximizes the I-score through step-wise elimination of variables from an initial subset sampled in some way from the variable space. The steps of the algorithm are as follows. Consider a training set $\{(y_1, x_1), ..., (y_n, x_n)\}$ of $n$ observations, where $x_i = (x_{1i}, ..., x_{pi})\}$ is a $p$-dimensional vector of explanatory variables. The size $p$ can be very large. All explanatory variables are discrete. Then we select an initial subset of $k$ explanatory variables $S_b = \{X_{b_1}, ..., X_{b_k}\}$, $b = 1, ..., B$. Next, we calculate $I(S_b) = \frac{1}{n\sigma^2}\sum_{j \in \mathcal{P}_k} n_j^2 (\bar{Y}_j - \bar{Y})^2$. For the rest of the paper, we refer this formula as Influence Measure or Influence Score (I-score). Tentatively drop each variable in $S_b$ and recalculate the I-score with one variable less. Then drop the one that gives the highest I-score. Call this new subset $S_b'$ which has one variable less than $S_b$. Continue to the next round of dropping variables in $S_b'$ until only one variable is left. Keep the subset that yields the highest I-score in the entire process. Refer to this subset as the \emph{return set} $R_b$. This will be most important and influential variable module from this initial training set. The above steps can be summarized in the following, see Algorithm \ref{alg:BDA}.

\begin{algorithm}[H]
\SetAlgoLined
\caption{\textbf{BDA.} Procedure of the Backward Dropping Algorithm (BDA) \label{alg:BDA}}
\emph{Training Set}: Consider a training set $\{(y_1, x_1), ..., (y_n, x_n)\}$ of $n$ observations, where $x_i = (x_{1i}, ..., x_{pi})\}$ is a $p$-dimensional vector of explanatory variables. The size $p$ can be very large. All explanatory variables are discrete.\;
\emph{Sampling from Variable Space}: Select an initial subset of $k$ explanatory variables $S_b = \{X_{b_1}, ..., X_{b_k}\}$, $b = 1, ..., B$\;
    \emph{Initialization}: Set $l = k$\;
    \While{While $l > 1$}{
        \emph{Compute Standardized I-score}: calculate $I(S_b) = \frac{1}{n\sigma^2}\sum_{j \in \mathcal{P}_k} n_j^2 (\bar{Y}_j - \bar{Y})^2$. For the rest of the paper, we refer this formula as Influence Measure or Influence Score (I-score). \\
        \emph{Drop Variables}: Tentatively drop each variable in $S_b$ and recalculate the I-score with one variable less. Then drop the one that gives the highest I-score. Call this new subset $S_b'$ which has one variable less than $S_b$. \\
        l = $|S_b'|$ (update $l$ with length of current subset of variables)
    }
\end{algorithm}

\subsection{A Toy Example}
\label{subsec:toy-example}

Let us create the following simulation to further investigate the advantage the proposed I-score has over the conventional measure AUC. In this experiment, suppose there are $X_1, ..., X_{10} \sim_\text{iid} \text{Bernoulli}(0.5)$. In other words, the sequence of all 10 independent variables defined above can only take values from $\{1, 0\}$. We create a sample of 2,000 observations for this experiment. Suppose we define the following model
\begin{equation}
    Y = X_1 + X_2 (\text{mod } 2)
\end{equation}
while ``mod'' refers to modulo of 2, i.e. $1 + 1 = 2 \equiv 0$. We can compute the AUC values and the I-score values for all 10 variables. In addition, we also compute both measures for the following models (this is to assume when we do not know the true form of the real model): (i) $X_1 + X_2$, (ii) $X_1 - X_2$, (ii) $X_1 \cdot X_2$, (iv) $X_1 / (X_2 + \epsilon)$. In model (iv), we add $\epsilon = 10^{-5}$ to ensure the division is legal. To further illustrate the power of I-score statistics, we introduce a new variable specifically constructed by taking the advantage of partition retention, i.e. $X^\dagger := \bar{y}_j$ while $j \in \Pi_{X_1,X_2}$ (the novel dagger technique that is defined using variable partition is widely used in application, see equation \ref{eq:interaction-based-features}). The simulation has 2,000 observations. We make a 50-50 split. The first 1,000 observations are used to create partitions and the local average of the target variable $Y$ that is required in creating $X^\dagger$ feature is only taken from the first 1,000 observations. In the next 1,000 observations, we can directly observe $\{X_1, X_2\}$ and retain the partition. For each of the partition, we then go to training set (the first 1,000 observations) and use the $\bar{y}_j$ values created using only the training set. In other words, the  We present the simulation results in Table \ref{tab:sim-result-one-module}.

The simulation results show that the proposed I-score is capable of selecting truly predictive variables while the AUC values might be misleading. We can see that the challenge of this toy dataset is that the variables do not have marginal signal. In other words, variables alone do not have predictive power. We can see this from the average AUC and I-score values. Since AUC value cannot be computed using multiple variables directly, assumptions of the underlying models are built. We guess four models: (i) $X_1 + X_2$, (ii) $X_1 - X_2$, (ii) $X_1 \cdot X_2$, (iv) $X_1 / (X_2 + \epsilon)$ and these models all have low AUC values even though the variables are correct. This implies under false assumption of the underlying model, AUC produce no reliable measure of detecting how significant the variables are. However, the proposed statistics on the assumed models (i)-(iv) are much higher than on individual variables alone. This means that the proposed I-score has the capability of detecting important variables even under incorrect assumption of the model formulation. 

The section ``Guessed'' in Table \ref{tab:sim-result-one-module} consists of 4 assumed models containing the correct variables, $X^\dagger$, and the variable set $\{X_1, X_2\}$. In this section, we observe that the AUC values are only high for $X^\dagger$ and are rather poor in the rest of assumed model. In addition, we cannot compute AUC for a variable set $\{X_1, X_2\}$. Alternatively, we can use I-score. The I-score values are drastically different in this section than the rest of the paper. The I-score for model (i) and (ii) are both above 700. The I-score for $X^\dagger$ is exactly the same as the theoretical value and so is the AUC for $X^\dagger$. This means the $X^\dagger$ technique takes the advantage of partition of the variable set successfully contains all the information that we need in order to make the best prediction result. This is not yet discovered in the literature.

As a summary for this simulation, we can conclude the following
\begin{itemize}
    \item In the scenario when the data set has many noisy variables and each variable observed does not have any marginal signal, the common practice AUC value will miss the information. This is because AUC still relies on the marginal signal. In addition, AUC is defined under the response variable $Y$ and its predictor $\hat{Y}$ which requires us to make assumption on the underlying model formulation. This is problematic because the mistakes carried over in making the assumption can largely affect the outcome of AUC. However, this challenge is not a roadblock for the proposed statistics I-score at all. In the same scenario with no marginal signal, as long as the important variables are involved in the selection, I-score has no problem signaling us the high predictive power disregard whether correct form of the underlying model can be found or not.
    \item The proposed I-score is defined using the partition of the variable set. This variable set can have multiple variables in it and the computation of I-score does not require any assumption of the underlying model. This means the proposed I-score does not subject to the mistakes carried over in the assumption or searching of the true model. Hence, I-score is a non-parametric measure.
    \item The construction of I-score can also be used to create a new variable that is based on the partition of any variable set. We call this new variable $X^\dagger$, hence the name ``dagger technique''. It is a engineering technique in our work that combines a variable set to form a new variable that contains all the predictive power that the entire variable set can provide. \textbf{This is a very powerful technique due to its high flexibility. In addition, it can be constructed using the variables with high I-score value after Backward Dropping Algorithm.}
\end{itemize}

\begin{table}
    \centering
    \caption{\textbf{Simulation Results.} This table presents the simulation results for the model $Y = X_1 + X_2 (\text{mod } 2)$. In this simulation, we create a toy data with just 10 variables (all drawn from Bernoulli distribution with probability $1/2$). We define the true model using only the first two variables and the remaining variables are noisy information. The task is to present the results of the AUC and the I-score values. The experiment is repeated 30 times and we present the average and standard derivation (SD) of the AUC and I-score values. We can see that there is no marginal information contributed by any of the variables alone, because each variable by itself has low AUC values and I-score values below 1 (if I-score is below 1, this provide almost no predictive power). We can use the guessed model (i) to (iv) that is composed of the true variables (here we assume that we know $\{X_1, X_2\}$ is important but we do not know the true form). We assign $\epsilon = 0.0001$ to ensure the division in model (iv) to be legal in case $X_2 = 0$. Last, we present the true model as a benchmark. Note: the ``NA'' entry means that AUC cannot be computed, i.e. not applicable or NA. \textbf{The measure of AUC values has a major drawback: it cannot successfully detect the useful information. Even with the correct variables selected (all guessed models only use the important variables $\{X_1, X_2\}$), AUC measure subjects to serious attack from incorrect model assumption. This flaw renders applications of using AUC measure to select models less ideal and sub-optimal. However, the proposed I-score is capable of indicating the most important variables, $X_1$ and $X_2$, disregard the forms of the underlying model. Moreover, the dagger technique of building $X^\dagger$ using partitions generated by the variable set $X_1$ and $X_2$ completely recovers full information of the true model even before any machine learning or model selection procedure, which is a novel invention that the literature has not yet seen.}
    }
    \resizebox{\textwidth}{!}{
        \begin{tabular}{cccccc}
        \toprule
        	&		&	Average AUC	&	SD. of AUC	&	Average I-score	&	SD. of I-score	\\
        \hline
        \multirow{2}{2.2cm}{Important}	&	$X_1$	&	0.51	&	0.01	&	0.49	&	0.52	\\
        	&	$X_2$	&	0.51	&	0.01	&	0.52	&	0.68	\\
        \hline
        \multirow{8}{2.2cm}{Noisy}	&	$X_3$	&	0.51	&	0.01	&	0.65	&	0.71	\\
        	&	$X_4$	&	0.5	&	0.01	&	0.41	&	0.6	\\
        	&	$X_5$	&	0.51	&	0.01	&	0.61	&	0.77	\\
        	&	$X_6$	&	0.5	&	0.01	&	0.27	&	0.29	\\
        	&	$X_7$	&	0.51	&	0.01	&	0.42	&	0.7	\\
        	&	$X_8$	&	0.5	&	0.01	&	0.33	&	0.48	\\
        	&	$X_9$	&	0.51	&	0.01	&	0.49	&	0.68	\\
        	&	$X_{10}$	&	0.51	&	0.01	&	0.39	&	0.48	\\
        \hline
        \multirow{4}{2.2cm}{Guessed models (using $\{X_1,X_2\}$)}	&	model (i): $X_1+X_2$	&	0.51	&	0.01	&	749.68	&	0.61	\\
        	&	model (ii): $X_1 - X_2$	&	0.51	&	0.01	&	749.54	&	0.37	\\
        	&	model (iii): $X_1 \cdot X_2$	&	0.49	&	0.25	&	248.2	&	18.26	\\
        	&	model (iv): $X_1/(X_2+\epsilon)$	&	0.57	&	0.11	&	250.03	&	11.72	\\
        	&	$X^\dagger$ (see eq. \ref{eq:interaction-based-features})	&	1	    &	0	    &	999.14	&	0.54	\\
        	&   $\{X_1, X_2\}$ &  NA    &    NA     &   500.08  &   0.48 \\
        \hline
        \multirow{1}{2.2cm}{True model}	&	$X_1 + X_2 \text{ (mod 2)}$	&	1	&	0	&	999.14	&	0.54	\\
        \bottomrule
        \end{tabular}
    }
    \label{tab:sim-result-one-module}
\end{table}

\subsection{Why I-score?}
\label{subsec:why-iscore}

\paragraph{Why is I-score the best candidate?}{
    The design of the proposed I-score has the following three benefits. First, I-score is a non-parametric measurement. This means I-score does not require any model fitting procedure. Second, the behavior of I-score is parallel with that of AUC. If a predictor has high I-score, this predictor must have high AUC value. However, AUC cannot be directly defined if we have a set of variables. In this case, an estimation of the true model must be used in order to compute AUC which means any attack from incorrect estimation of the true model will lower AUC value. However, I-score does not subject to this attack. Third, the proposed I-score can be developed into an interaction-based feature engineer method called the ``dagger technique''. This technique can recover all useful information from a subset of variables. Next, we will explain each of the above three reasoning. 
}

\paragraph{Non-parametric Nature}{
    The proposed I-score (equation \ref{eq:iscore}) does not rely or make any assumptions on the model fitting procedure. The procedure of model fitting refers to the step of searching for a model that estimate the true relationship between features and target variable. Suppose we have a subset of the features $\textbf{X} \in \mathbb{R}^d$ and their corresponding target variable $Y$. The relationship between $\textbf{X}$ and $Y$ (denote this relationship to be $f(\cdot)$) is often times not observable. We would have to make an assumption of the underlying model $\hat{f}(\cdot)$ in order to construct a predictor $\hat{Y}:= \hat{f}(\textbf{X})$. Only then can we measure the difference between our estimate and the true target variable. We can do so by using a loss function $\mathcal{L}(Y, \hat{Y})$ to estimate the error the predictor $\hat{Y}$ has at predicting the target variable $Y$. Disregard the numerical magnitude of the result from the loss function, we are uncertain the source of the mistakes we have. This is because the predictor $\hat{Y}$ is composition of both the estimation of the true model $\hat{f}(\cdot)$ and the selected features $\textbf{X}$. The literature has not yet discovered any procedure to directly measure the influence $\textbf{X}$ has on target variable $Y$ without the estimation of true model $\hat{f}(\cdot)$. Hence, the error, measured by loss $\mathcal{L}(Y, \hat{Y})$ or ROC AUC values, is confounded and we do not know clearly how the features $\textbf{X}$ influence the target variable $Y$ if it has any influence.
    
    The proposed I-score can directly measure the influence of features $\textbf{X}$ has on target variable $Y$ without the usage of any estimation of the true model. In other words, the formula of I-score (equation \ref{eq:iscore}) has no component of $\hat{f}(\cdot)$ and there is no need to have $\hat{f}(\cdot)$ as long as we can compute I-score. This clears our judgement of feature selection and allows the end-users to immediately know not only the impact of the features $\textbf{X}$ but also how important they are at predicting target variable $Y$.
    
    From the toy example (see the simulation results in Table \ref{tab:sim-result-one-module} in subsection \ref{subsec:toy-example}), we can see that the estimation of the true model, if incorrect, produce extremely poor AUC values (model (i) - (iv) in Table \ref{tab:sim-result-one-module} have AUC values approximately 50\%, equivalent to random guessing) even though the important and significant features are present in the variable set. I-score does not subject to the attack of incorrect estimation of the true model. We can see that I-score of the same incorrect models are extremely high as long as the important features, $\{X_1,X_2\}$, are included and present. Consequently, we regard this as the most important contribution and reason why I-score and the ``dagger technique'' are the best tools.
}

\paragraph{High I-score Produces High AUC Values}{
    We provide in the subsection \ref{subsec:iscore-confusion-auc} a technical analysis on why high I-score induce high AUC values. In the anatomy of I-score formula in subsection \ref{subsec:iscore-confusion-auc}, we can see that the first partition (this is the partition when $\hat{Y}$ is positive) is directly related to sensitivity. In addition, we reward this outcome with a factor of $n_1^2$ where $n_1$ is the number of observations that falls in this partition, i.e. $\hat{Y}$ is positive. This construction raises I-score significantly and proportionate to the number of true positive cases which is a major component causing AUC value to rise. Hence, the benefit of using I-score is that it allows statisticians and computer scientists to immediately identify the features that can raise the AUC values. This parallel behavior allows easy interpretation of I-score values for its end-users. 
    
    Researchers have pointed out that ROC treats sensitivity and specificity equally \cite{halligan2015disadvantages} which can cause lack of interpretability in some areas of practice such that the diagnostic is focusing on true positives. Poor sensitivity could mean that the experiment is missing the positive cases. In other words, AUC can consider  a test that increase sensitivity at low specificity superior to one that increases sensitivity at high specificity. It is recommended that better tests should reward sensitivity to avoid false positives \cite{baker2003central}. The construction of the I-score directly rewards sensitivity proportionate to the number of observations classified positive cases correctly which allows I-score to exponentially grow as the sample size in training set increases. This nature allows I-score to not only behave in parallel to AUC values but is capable of reflecting the sensitivity values proportionate to the sample size. 
}

\paragraph{I-score and the ``Dagger Technique''}{
    The ``dagger technique'' is the most innovative method ever introduced based on the literature in developing I-score \cite{lozheng2002} \cite{chernoffetal2009} \cite{lochernoffzhenglo2015} \cite{lochernoffzhenglo2016}. The proposed ``dagger technique'' organically combines the raw information from each feature by using partition retention \cite{chernoffetal2009} which allows the final combined feature to enrich the prediction performance free of incorrect estimation of the true model. This nature of the ``dagger technique'' is reflected in Table \ref{tab:sim-result-one-module}. The guessed models are all formulated using the correct variable sets: $\{X_1, X_2\}$. However, model (i) - (iv) produced shocking 50\% AUC values because AUC subjects to the attack of incorrect model estimation. This is a major drawback of using AUC to conduct model selection. However, the ``dagger technique'', as long as the variable set includes the important variables, produce an outstanding 100\% AUC with no variation. This means the ``dagger technique'' is capable of reconstructing the original information and enrich the predictor. We can see that the remaining methods all have some standard deviation when we simulate experiment with a different seed. However, the setting of different seed has no impact on the proposed ``dagger technique'' which means any time we use the ``dagger technique'' we are able to recover the full information of the target variable $Y$ with the correct variable set $\{X_1, X_2\}$. This fact justifies the remarkable value of the ``dagger technique'' towards the literature. The only other place that this is true in Table \ref{tab:sim-result-one-module} is the model. \textbf{Thus, the ``dagger technique'' has the capability to recover the full information in target variable if the features are included correctly and it does not subject to any attacks of false specification or estimation of the true model. We strongly recommend practitioners to use the proposed I-score and the ``dagger technique'' in the future whenever possible. }
}

\section{Application}
\label{sec:application}

\subsection{Language Modeling}
\label{subsec:language-modeling}

\paragraph{N-gram}{
    In Natural Language Processing, the prediction of the probability of a word or the classification of the label of a paragraph is a fundamental goal in building language models. The most basic language model is the ``N-gram'' models.
    
    In ``N-gram'' modeling, a single word such as ``machine'' as a feature is a uni-gram ($n=1$). The 2-word phrase ``machine learning'' is a bi-gram ($n=2$). The 3-word phrase ``natural language processing'' is a tri-gram ($n=3$). Any text document can be processed to construct N-gram features using this technique. For example, we may observe a sentence and it states ``I love this movie''. If the goal is to classify the sentence into one of the two classes: $\{\text{positive}, \text{negative}\}$, a common practice is to estimate the probability of the label $Y$ belongs to a certain class based on the previous words in the same sentence:
    \begin{equation}
        \label{eq:example-N-gram}
        \begin{array}{l}
            \mathbb{P}(Y=1|\text{i}), \text{ uni-gram} \\
            \mathbb{P}(Y=1|\text{i love}), \text{ bi-gram} \\
            \mathbb{P}(Y=1|\text{i love this}), \text{ tri-gram} \\
            \mathbb{P}(Y=1|\text{i love this movie}), \text{ 4-gram} \\
        \end{array}
    \end{equation}
    
    Joulin et. al. (2016) \cite{joulin2016bag} summarized a list of NLP algorithms to produce a range of performance from 80\% to 90\% range. They also show evidence that N-gram models can increase the accuracy slightly more by using more N-grams. However, the aggregate N-gram models are not yet discovered. In addition, continuing adding N-gram features would result in overfitting which leads to more and more complex architecture of RNN such as Long Short Term Memory (LSTM), Gated Recurrent Unit (GRU), and bi-directional RNN. \textbf{Thus far, there has not been any dimensionality reduction technique implemented within a RNN in the literature.}
}

\begin{figure}
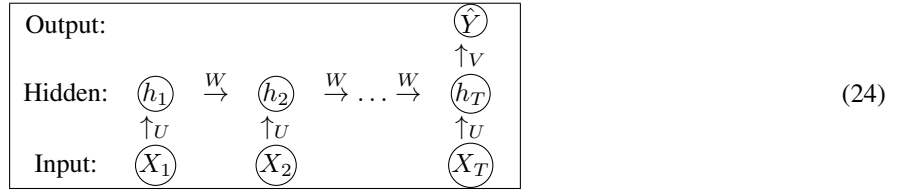

    \centering
    \caption{\textbf{A Simple RNN for Text Classification.} This diagram illustrates the basic steps of using RNN for text classification. The input features are $\{X_1, X_2, ...\}$. The hidden neurons are $\{h_1, h_2, ...\}$. The output prediction is $\hat{Y}$. Since this is the text classification problem, the architecture has many inputs and one output, hence the name ``many-to-one''. The architecture has parameters $\{U, V, W\}$ and these weights (or parameters) are shareable throughout the architecture. }
    \begin{equation}
    \begin{array}{|l|}
        \hline
        \begin{matrix}
            \text{Output:} & & & & & \circlesign{\hat{Y}} \\
             &  & & & & \uparrow_V \\
            \text{Hidden:} & \circlesign{h_1} & \stackrel{W}{\rightarrow} & \circlesign{h_2} & \stackrel{W}{\rightarrow} \dots \stackrel{W}{\rightarrow} & \circlesign{h_T} \\
             & \uparrow_U & & \uparrow_U & & \uparrow_U \\
            \text{Input:} & \circlesign{X_1} & & \circlesign{X_2} & & \circlesign{X_T}
        \end{matrix} \\
        \hline
    \end{array}
    \end{equation}
    \label{fig:basic-rnn}
\end{figure}

\paragraph{Recurrent Neural Network}{
    We introduced N-gram models above where the conditional probability of the label of a sentence $y$ only depends on the previous $n-1$ words. Let us briefly discuss the basic RNN that we will use in the application. The diagram for the basic RNN is presented in Figure \ref{fig:basic-rnn}. Suppose we have input features $X_1, X_2, ...$. These features are directly processed from the text document which can be processed word index or they can be embedded word vectors. The features are fed into the hidden layer where the neurons (or units) are denoted as $h_1, h_2, ...$. There is a weight connecting the previous neuron with the current neuron. We denote this weight as $W$. Each current neuron has contribution from current feature which is connected with a weight parameter $U$. For any $t$ in $\{1, 2, ..., T\}$, we can compute each hidden neuron by using the following formula
    \begin{equation}
        \label{eq:rnn-hidden-neuron}
        h_t = g(W \cdot h_{t-1} + U \cdot X_t + b)
    \end{equation}
    where $W$ and $U$ are trainable parameters, $b$ is the bias term, and $g(\cdot)$ can be an activation function. This choice of the activation function is completely determined by the dataset and the end-user. A list of famous activation functions can be found in Table \ref{tab:activationfct}. In the end of the architecture, we can finally compute the predicted probability of $Y$ given the hidden neurons by using the formula
    \begin{equation}
        \label{eq:rnn-output}
        \hat{Y} = g(V \cdot h_T + b)
    \end{equation}
    where the weights $W$, $U$, and $V$ are shareable in the entire architecture. Forward propagation of RNN is referring to the procedure when information flow from the input features to output predictor using equation \ref{eq:rnn-hidden-neuron} and equation \ref{eq:rnn-output}.
}

\paragraph{Backward Propagation Using Gradient Descent}{
    The forward propagation allows us to pass information from input layer which are features extracted from text document to the output layer which is the predictor. To search for the optimal weights, we need to compute the loss function and optimize the loss function by updating the weights. Since the task is text classification, we only have one output in the output layer. In addition, the task is a two-class classification problem because $Y$ can only take value of $\{0, 1\}$. This means we can define the loss function using cross-entropy function, which is defined as 
    \begin{equation}
        \label{eq:cross-entropy}
        \mathcal{L}(Y, \hat{Y}) = - \sum_{i=1}^n y_i \log \hat{y}_i - (1 - y_i) \log (1 - \hat{y}_i)
    \end{equation}
    where $y_i$ is the true label for instance $i$ and $\hat{y}_i$ is the predicted value for instance $i$. Notice that in the proposed architecture we use I-score and the ``dagger technique'' to construct $\text{X}^\dagger$ for the input features where $T'$ is smaller than $T$ and it is a length dependent on tuning process. Next, we can use gradient descent (GD) to update the parameters. At each step $s$, we can update the weights by 
    \begin{equation}
        \label{eq:gd-update-weights}
        \begin{array}{rcl}
            V_s &=& V_{s-1} - \eta \cdot \nabla_V \mathcal{L}(Y, \hat{Y}) \\
            U_s &=& U_{s-1} - \eta \cdot \nabla_U \mathcal{L}(Y, \hat{Y}) \\
            W_s &=& W_{s-1} - \eta \cdot \nabla_W \mathcal{L}(Y, \hat{Y}) \\
        \end{array}
    \end{equation}
    where $\eta$ is learning rate (usually a very small number), the symbol $\nabla$ means gradient, and $\nabla \mathcal{L}(\cdot)$ is the gradient (partial derivative) of the loss function $\mathcal{L}(\cdot)$. We can formally write the gradients in the following
    \begin{equation}
        \label{eq:gd-gradients}
        \begin{array}{rcl}
            \nabla_V \mathcal{L}(Y, \hat{Y}) &=& \frac{\partial}{\partial V} \mathcal{L}(Y, \hat{Y}) \\
            \nabla_U \mathcal{L}(Y, \hat{Y}) &=& \frac{\partial}{\partial U} \mathcal{L}(Y, \hat{Y}) \\
            \nabla_W \mathcal{L}(Y, \hat{Y}) &=& \frac{\partial}{\partial W} \mathcal{L}(Y, \hat{Y})
        \end{array}
    \end{equation}
    where $\nabla_\text{parameter}$ means gradient or partial derivative with respect to that parameter.
}

\begin{figure}
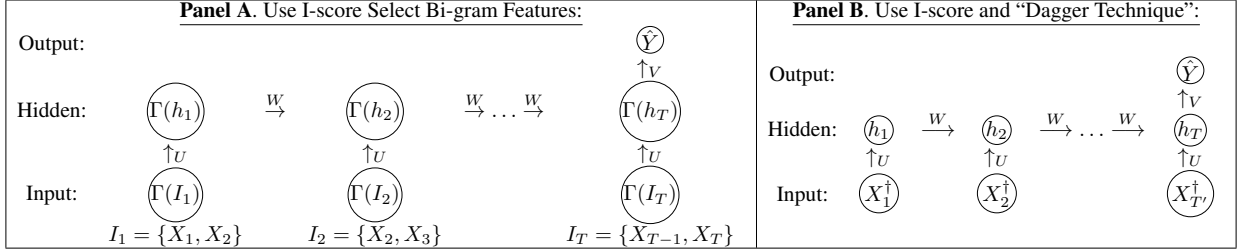

    \centering
    \caption{\textbf{Executive Diagram for Proposed Method.} This figure represents the proposed methodologies using N-grams, I-score, and the ``dagger technique''. The forward propagation and the backward propagation remains the same before and after using I-score. The function $\Gamma(\cdot)$ acts as a gate to release the neuron according to certain I-score criteria. (A) Panel A presents Recurrent Neural Network design to implement I-score at each step of the neural network architecture. (B) Panel B presents features constructed using ``dagger techniqe'' and then fed into the Recurrent Neural Network (RNNs) architecutre. }
    \resizebox{\textwidth}{!}{
        $
            \begin{array}{| c | c |}
            \hline
            \text{\underline{\textbf{Panel A}. Use I-score Select Bi-gram Features:}} & \text{\underline{\textbf{Panel B}. Use I-score and ``Dagger Technique'':}} \\
            \begin{matrix}
                \text{Output:} & & & & & \circlesign{\hat{Y}} \\
                 &  & & & & \uparrow_V \\
                \text{Hidden:} & \circlesign{\Gamma(h_1)} & \stackrel{W}{\rightarrow} & \circlesign{\Gamma(h_2)} & \stackrel{W}{\rightarrow} \dots \stackrel{W}{\rightarrow} & \circlesign{\Gamma(h_T)} \\
                 & \uparrow_U & & \uparrow_U & & \uparrow_U \\
                \text{Input:} & \circlesign{\Gamma(I_1)} & & \circlesign{\Gamma(I_2)} & & \circlesign{\Gamma(I_T)} \\
                 & I_1 = \{X_1, X_2\} & & I_2 = \{X_2, X_3\} & & I_T = \{X_{T-1}, X_T\}
            \end{matrix}
            &
            \begin{matrix}
                \text{Output:} & & & & & \circlesign{\hat{Y}} \\
                 &  & & & & \uparrow_V \\
                \text{Hidden:} & \circlesign{h_1} & \stackrel{W}{\longrightarrow} & \circlesign{h_2} & \stackrel{W}{\longrightarrow} \dots \stackrel{W}{\longrightarrow} & \circlesign{h_T} \\
                 & \uparrow_U & & \uparrow_U & & \uparrow_U \\
                \text{Input:} & \circlesign{X^\dagger_1} & & \circlesign{X^\dagger_2} & & \circlesign{X^\dagger_{T'}}
            \end{matrix} \\
            \hline
        \end{array}
        $
    }

    \label{fig:executive-diagram} 
\end{figure}

\paragraph{Implementation with I-score}{
    This section we introduce the proposed algorithm which is built upon the idea of a novel statistical measure I-score. The executive diagram for ``N-gram'' models using I-score as feature selection and engineering technique is presented in Figure \ref{fig:executive-diagram}.
    
    Figure \ref{fig:executive-diagram} has presented two main pipelines for this paper that implements the proposed I-score into a basic RNN structure. Panel A presents the first simple framework of combining ``N-grams'' with I-score. The procedure of ``N-grams'' process text data into averaged numerical data. The conventional setup is presented in the left plot in Figure \ref{fig:basic-rnn}. The text document is processed using text vectorization and then it is embedded into a matrix form. We can then fit a basic feed-forward ANN or sequential RNN. Panel B presents an alternative input using the ``dagger technique''. We introduce them as follows.
    
    The proposed I-score can be used in the following approaches:
    \begin{itemize}
        \item First, we can compute I-score for each RNN unit. For example, in the Panel A of Figure \ref{fig:executive-diagram}, we can first compute I-score on text vectorization layer. Then we can compute I-score on the embedding layer. With the distribution of I-score provided from the feature matrix, we can use a particular threshold to identify the cutoff used to screen for important features that we need to feed into the RNN architecture. We denote this action by using $\Gamma(\cdot)$ and it is defined as $\Gamma(\mathcal{X}) := \mathcal{X} \cdot \mathds{1}(\text{I}(Y, \mathcal{X}) > \text{threshold})$. For input layer, each feature $I_t$ can be released or omitted according to its I-score values. That is, we use $\Gamma(I_t) := I_t \cdot \mathds{1}(\text{I}(Y, I_t) > \text{threshold})$ to determine whether this input feature $I_t$ is predictive and important enough to be fed into the RNN architecture. For hidden layer, each hidden neuron $h_t$ can be released or omitted according to its I-score values. In other words, we can use $\Gamma(h_t) := h_t \cdot \mathds{1}(\text{I}(Y, h_t) > \text{threshold})$ to determine whether this hidden neuron $h_t$ is important enough to be inserted in the RNN architecture. If at certain $t$ the input feature $X_t$ fails to meet certain I-score threshold ($X_t$ would fail if the I-score of $X_t$ is too low, then $\Gamma(X_t) = 0$), then this feature is not fed into the architecture and the next unit $h_{t+1}$ is defined, using equation \ref{eq:rnn-hidden-neuron}, as $h_{t+1} = g(W \cdot h_{t-1} + U \cdot 0 + b)$. This is the same for any hidden neuron as well. If certain hidden neuron $h_t$ has the previous hidden neuron $h_{t-1}$ fails to meet I-score criteria, then $h_t$ is defined as $h_t = g(W \cdot 0 + U \cdot X_t + b)$. \textbf{Hence, this $\Gamma(\cdot)$ function acts as a gate to allow the information of the neuron $h_t$ to pass through according to a certain I-score threshold. If $\Gamma(I_t)$ is zero, that means this input feature is not important at all and hence can be omitted by replacing it with zero value. In other words, it is as if this feature never existed. In this case, there is no need to construct $\Gamma(h_t)$. We show later in section \ref{sec:application} that important long-term dependencies that are associated with language semantics can be detected using this $\Gamma(\cdot)$ function, because I-score has the power to omit noisy and redundant features in the RNN architecture. Since I-score is compared throughout the entire length of $T$, the long-term dependencies between features that are far apart can be captured using high I-score values.}
        
        \item Second, we can use the ``dagger technique'' to engineer and craft novel features using equation \ref{eq:interaction-based-features}. We can then calculate I-score on these dagger feature values to see how important they are. We can directly use 2-gram model and I-score is capable of indicating which 2-gram phrases are important. These phrases can act as two-way interactions. According to I-score, we can then determine whether we want all the words in the 2-gram models, 3-gram models, or even higher level of N-gram models. When $n$ is large, we recommend to use the proposed Backward Dropping Algorithm to reduce dimension within the N-word phrase before creating new feature using the proposed ``dagger technique''. For example, suppose we use 2-gram model. A sentence such as ``I love my cats'' can be processed into (I, love), (love, my), (my, cats). Each feature set has two words. We can denote the original sentence ``I love my cats'' into 4 features $\{X_1, X_2, X_3, X_4\}$. The ``dagger technique'' suggests that we can use equation \ref{eq:interaction-based-features} with 2-gram models fed in as inputs. In other words, we can take $\{X_1, X_2\}$ and construct $\bar{Y}_j$ where $j$ is the running index tracking the partitions formed using $\{X_1, X_2\}$. If we discretize $X_1$ and $X_2$ (see subsection \ref{subsec:discretization} for detailed discussion of discretization using I-score) and they both take values $\{0, 1\}$, then there are $2^2 = 4$ partitions and hence $j$ can take values $\{1, 2, 3, 4\}$. In this case, the novel feature $X^\dagger_1$ can take on 4 values, i.e. an example can be seen in Table \ref{tab:dagger-for-small-example}. \textbf{The combination of I-score, Backward Dropping Algorithm Algorithm, and the ``dagger technique'' allow us to prune the useful and predictive information in a feature set so that we can achieve maximum prediction power with as little number of features possible.}
        
        \item Third, we can concatenate many N-gram models with different $n$ values. For example, we can carry out N-gram modeling using $n=2$, $n=3$, and $n=4$. This way we can more combination of higher order interactions. In order to avoid overfitting, we can use I-score to select the important interactions and then use these selected phrases (which can be two-word, three-word, or four-word) to build RNN models.
    \end{itemize}
}

\subsection{IMDB Dataset}
\label{subsec:imdb-data}

In this application, we use the IMDB Movie Database which includes 25,000 paragraphs in training and testing set. In total, there are 50,000 paragraphs and each paragraph carries a label. The label is dichotomous, i.e. $Y=1$ if the review is positive and $Y=0$ if the review is negative. A sample of the data is presented in Table \ref{tab:subset-of-data}. The first sample is a positive review. The second sample is a negative review.

In the experiment of text classification, we use data provided in IMDB move database. The dataset consists of 25,000 observations in training set and testing set each. In total, there are 50,000 observations. The data set has equal portion for two classes. The two classes are $Y=1$ if the movie review is positive and $Y=0$ if the movie review is negative. The goal is to read in a paragraph and predict whether the tone of this paragraph of movie review is positive or negative. We present sample data in the Table \ref{tab:subset-of-data}.

\begin{table}
    \centering
    \caption{\textbf{Sample Data and Selected Semantics.} This table presents sample data. We present two samples. The first column is the paragraphs directly taken from the IMDB movie database. The second column is the corresponding label for the first column. \textbf{The proposed I-score selects words that have significant association with the semantics of the sentence. This is because I-score selects features that are highly predictive of the target variable. In this application, the target variable carries tones and preferences of a movie of which the same writer provides critics and reviews. The semantics in the critics and reviews reflect the tones and the preferences of the movie reviewers which is why I-score is able to detect the features using the provided preferences in the label.} }
    \resizebox{\textwidth}{!}{
        \begin{tabular}{l p{9cm} p{5cm} p{6cm} c}
            \toprule
            \multirow{2}{*}{No.} & \multirow{2}{*}{Samples}	& \multicolumn{2}{c}{I-score Features} &	\multirow{2}{*}{Label} \\
            \cline{3-4}
            & & Uni-gram & 2-gram, 3-gram & \\
            \hline
            1 & $<\text{UNK}>$ this film was just brilliant casting location scenery story direction everyone's really suited the part they played and you could just imagine being there robert $<\text{UNK}>$ is an amazing actor and now the same being director $<\text{UNK}>$ father came from the same scottish island as myself so i loved the fact there was a real connection with this film the witty remarks throughout the film were great it was just brilliant so much that i bought the film as soon as it was released for retail and ...
            & \{congratulations, lovely, true\}
            & \{amazing actor, really suited\}
            &	1	\\
            2 & $<\text{UNK}>$ big hair big boobs bad music and a giant safety pin these are the words to best describe this terrible movie i love cheesy horror movies and i've seen hundreds but this had got to be on of the worst ever made the plot is paper thin and ridiculous the acting is an abomination the script is completely laughable the best is the end showdown with the cop and how he worked out who the killer is it's just so damn terribly written ...
            & \{bad, ridiculous\}
            & \{bad music, terribly written\}, \{damn terribly written\}
            &	0	\\
            \bottomrule
        \end{tabular}
    }
    \label{tab:subset-of-data}
\end{table}

In Table \ref{tab:subset-of-data}, we present evidence that I-score can detect important words that is highly impactful for predicting the label of positive or negative tones. In Table \ref{tab:iscore-semantics}, we present the semantics of the same two samples that I-score narrows down. The original RNN uses the processed word vector to make predictions. For example, scholars have been using N-gram models to extract features from the text document \cite{joulin2016bag}. Other common techniques of feature extractions are Term Frequency-Inverse Document
Frequency (TF-IDF), Term Frequency (TF) \cite{salton1988term}, Word2Vec \cite{goldberg2014word2vec}, and Global Vectors for Word
Representation (GloVe) \cite{pennington2014glove}. However, I-score is able to shrink the word counts with the reduced dimensions to be a tuning parameter. In this experiment, we reduce the number of words from 400 to 100 or even 30 while remaining the same semantics. \textbf{This dimension reduction technique is completely novel and can provide what goes beyond human intuition in language semantics problem in NLP.}

\begin{table}
    \centering
    \caption{\textbf{Interpreted Semantics Using I-score.} This table presents sample data. We present two samples. The first column is the paragraphs directly taken from the IMDB movie database. The second column presents features selected by I-score according to different I-score threshold (we use top 7.5\% and top 25\% as examples). The last column presents the corresponding label. \textbf{The semantics of the selected features are subset of words from the original sample. We observe that I-score can select subset of words while maintaining the same semantics.}  }
    \resizebox{\textwidth}{!}{
    \begin{tabular}{l p{8cm} p{3cm} p{6cm} c}
        \toprule
        \multirow{2}{*}{No.} & \multirow{2}{*}{Samples (Original Paragraphs)}	& \multicolumn{2}{c}{I-score Features (using different thresholds)} &	\multirow{2}{*}{Label} \\
        \cline{3-4}
        & & Top 7.5\% I-score & Top 25\% I-score & \\
        \hline
        1 & $<\text{UNK}>$ this film was just brilliant casting location scenery story direction everyone's really suited the part they played and you could just imagine being there robert $<\text{UNK}>$ is an amazing actor and now the same being director $<\text{UNK}>$ father came from the same scottish island as myself so i loved the fact there was a real connection with this film the witty remarks throughout the film were great it was just brilliant so much that i bought the film as soon as it was released for retail and would recommend it to everyone to watch and the fly fishing was amazing really cried at the end it was so sad and you know what they say if you cry at a film it must have been good and this definitely was also congratulations to the two little boy's that played the $<\text{UNK}>$ of ... 
        & \{congratulations often the play them all a are and should have done you think the lovely because it was true and someone's life after all that was shared with us all\}
        & \{for it really at the so sad you what they at a must good this was also congratulations to two little boy's that played the <UNK> of norman and paul they were just brilliant children are often left out the praising list i think because the stars that play them all grown up are such a big profile for the whole film but these children are amazing and should be praised for what they have done don't you think the whole story was so lovely because it was true and was someone's life after all that was shared with us all\}
        &	1	\\
        & 400 words & 31 words & 101 words \\
        \hline
        2 & $<\text{UNK}>$ big hair big boobs bad music and a giant safety pin these are the words to best describe this terrible movie i love cheesy horror movies and i've seen hundreds but this had got to be on of the worst ever made the plot is paper thin and ridiculous the acting is an abomination the script is completely laughable the best is the end showdown with the cop and how he worked out who the killer is it's just so damn terribly written the clothes are sickening and funny in equal measures the hair is big lots of boobs bounce men wear those cut tee shirts that show off their $<\text{UNK}>$ sickening that men actually wore them and the music is just $<\text{UNK}>$ trash that plays over and over again in almost every scene there is trashy music boobs and ... 
        & \{those $<\text{UNK}>$ every is trashy music away all aside this whose only is to look that was the 80's and have good old laugh at how bad everything was back then\} 
        & \{script best worked who the just so terribly the clothes in equal hair lots boobs men wear those cut shirts that show off their $<\text{UNK}>$ sickening that men actually wore them and the music is just $<\text{UNK}>$ trash that over and over again in almost every scene there is trashy music boobs and $<\text{UNK}>$ taking away bodies and the gym still doesn't close for $<\text{UNK}>$ all joking aside this is a truly bad film whose only charm is to look back on the disaster that was the 80's and have a good old laugh at how bad everything was back then\}
        &	0	\\
        & 400 words & 31 words & 101 words \\
        \bottomrule
    \end{tabular}
    }
    \label{tab:iscore-semantics}
\end{table}

\begin{figure}
    \centering
    \caption{\textbf{Learning Paths Before and After Discretization.} This figure presents the training procedure. All graphs present the training and validating paths. The first graph is from the original bi-gram data. The second is from using discretized bi-gram (discretized by I-score). The third is using the top 18 variables according to I-score values. \textbf{The proposed method can significantly improve computation efficiency.} }
    \includegraphics[width=.9\textwidth]{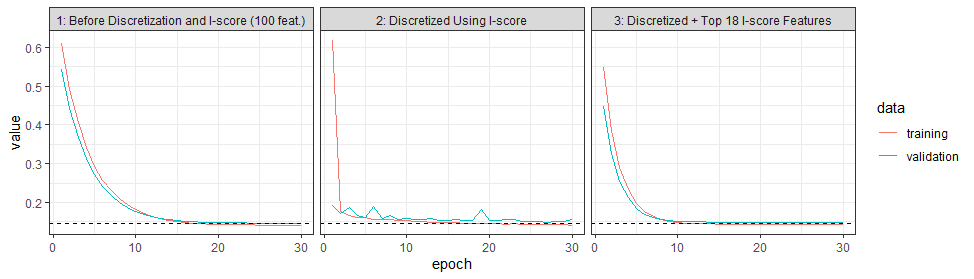}
    \label{fig:train_and_val_comparison}
\end{figure}

The training and validating process is summarized in Table \ref{fig:train_and_val_comparison}. In Figure \ref{fig:train_and_val_comparison}, we summarize the training and the validating paths that are generated from the experiment before and after using the proposed statistics I-score. The first plot of Figure \ref{fig:train_and_val_comparison} presents the training and validating paths for the original bi-gram data (here processed the data into 100 features). The second plot of Figure \ref{fig:train_and_val_comparison} is the same data but discretized using I-score. 

\begin{figure}
    \centering
    \caption{\textbf{Learning Paths Before and After Text Reduction Using I-score.} This figure presents the training procedure. All graphs present the training and validating paths. The first graph is from the original bi-gram data. The second is from using discretized bi-gram (discretized by I-score). The third is using the top 18 variables according to I-score values. \textbf{The proposed method can significantly improve computation efficiency.} }
    \includegraphics[width=.8\textwidth]{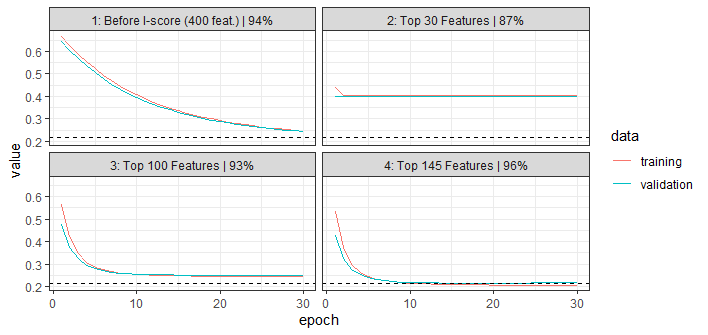}
    \label{fig:train_and_val_comparison_from400}
\end{figure}

Figure \ref{fig:train_and_val_comparison_from400} presents the learning paths before and after dimensionality reduction using the proposed I-score. To demonstrate the potential of I-score, we first use the extracted features from the embedding layer which generates 400 features (a tuning result). We can use these 400 features to send into a feed-forward ANN or sequential RNN. The test set performance is 94\% which is measured using AUC values. We can compute marginal I-score (this means we compute the I-score using each variable as a predictor independently). Amongst these 400 features, we can rank them using I-score values and pick the top 30 features. We feed these 30 features into a feed-forward ANN or sequential RNN and we already achieved 87\% on test set. To further improve the learning performance, we can increase the I-score threshold so that we can include more top influential features. We can use top 30, 100, and 145 respectively. We plot the learning paths in Figure \ref{fig:train_and_val_comparison_from400}. The first graph ``1'' is the learning path for the original 400 features. We can see that the training set and validating set error merely breached 0.2 in 30 epochs. However, when we use top 100 features, we are able to see in graph ``3'' that we are able to achieve near convergence of approximately 10 epochs. \textbf{This increased convergence speed is largely due to the nature that I-score is able to erase the noisy and redundant features from the input layer. In addition, I-score is able to deliver this efficient learning performance with only 25\% of the number of the original features which is something the literature has not yet seen. We regard this another major benefit of using the proposed technique in training neural networks.}

\subsection{Result}
\label{subsec:result}

We show in Table \ref{tab:imdb-results} the experiment results for the text classification task in IMDB Movie Dataset. We start with bi-gram models. The performance of using I-score to select the important bi-gram features produced 96.5\% AUC on the test set while the bi-gram model without I-score produced 92.2\%. We also used a combination of different N-gram models while we set the level of $N$ to be $\{2,3,4\}$, i.e. 2-gram, 3-gram, and 4-gram, respectively. We can concatenate the N-gram features and then feed the features into a feed-forward neural networks directly. First, we concatenate the 2-gram and 3-gram models together and then we use I-score to reduce the prediction performance. We recommend to use the top 5\% I-score threshold to screen for the important semantics among all the 2-gram and 3-gram features. This corresponds to approximately 40 out of the 800 concatenated 2-gram and 3-gram features. We observe that I-score is able to raise the prediction performance to 97.7\% while a combination of 2-gram and 3-gram model without I-score is only able to produce 91.1\% which is a 74\% error reduction. 

\begin{table}
    \centering
    \caption{\textbf{Test Performance Comparison.} The table presents experiment results of prediction performance on IMDB data set. The performance is measured in AUC and the proposed method increases the prediction result to 97\% which is 85\% error reduction.}
    \begin{tabular}{lc}
        \toprule
        Model & IMDB (Test Set) \\
        \hline
        CNN \cite{tang2015effective} & 37.5\% \\
        RNN & 85.7\% \\
        LSTM & 86.6\% \\
        GRU & 86.7\% \\
        2-gram & 92.2\% \\
        2-gram + 3-gram & 91.1\% \\
        2-gram + 3-gram + 4-gram & 90.2\% \\
        Average: & 81\% \\
        \hline 
        Proposed: \\
        2-gram: use high I-score features & 96.5\% \\
        2-gram + 3-gram: use high I-score features & 96.7\% \\
        2-gram + 3-gram + 4-gram: use high I-score features & 97.2\% \\
        2-gram + 3-gram + 4-gram: use novel ``dagger'' features (based on eq. \ref{eq:interaction-based-features}) & 97.2\% \\
        \bottomrule
    \end{tabular}
    \label{tab:imdb-results}
\end{table}

\section{Conclusion}
\label{sec:conclusion}
This paper proposes a novel I-score to detect and search for the important language semantics in text document that are useful for making good prediction in text classification tasks. 

\textbf{Theoretical Contribution.} We provide theoretical and mathematical reasoning why I-score can be considered as a function of AUC. The construction of I-score can be analyzed using partitions. We see from mathematical rearrangements of the I-score formula that sensitivity plays a major component. This role of I-score provides the fundamental driving force to raise AUC values if the variables selected to compute I-score are important and significant. In addition to its theoretical parallelism with AUC, I-score can be used anywhere in a neural network architecture which allows its end-users to flexibly deploy this computation, which is a nature that does not belong to AUC. AUC is also vulnerable under incorrect model specification. Any estimation of the true model, disregard whether it is accurate or not, is harmless to the performance of I-score due to its non-parametric nature, which is a novel measure for feature selection that the literature has not yet seen. 

\textbf{Backward Dropping Algorithm.} We also propose a greedy search algorithm called the Backward Dropping Algorithm that handles long-term dependencies in the dataset. Under the curse of dimensionality, the Backward Dropping Algorithm is capable of efficiently screening out the noisy and redundant information. The design of the Backward Dropping Algorithm also takes advantage of the nature of I-score, because I-score increases when the variable set has less noisy features while I-score decreases when the noisy features are included.

\textbf{Dagger Technique} We propose a novel engineering technique called the ``dagger technique'' that combines a set of features using partition retention to form a new feature that fully preserve the relationship between explanatory variable and response variable. This proposed ``dagger technique'' can successfully combine words and phrases with long term dependencies into one new feature that carries long term memory. It can also be used in constructing the features in many other types of deep neural networks such as Convolutional Neural Networks (CNNs). Though we present empirical evidence in sequential data application, this ``dagger technique'' can actually go beyond most image data and sequential data.

\textbf{Application.} We show with empirical results that this ``dagger technique'' can fully reconstruct target variable with the correct features, which is a method that can be generalized into any feed-forward Artificial Neural Networks (ANNs) and Convolutional Neural Networks (CNNs). We demonstrate the usage of I-score and proposed methods with simulation data. We also show with a real world application on the IMDB Movie Dataset that the proposed methods can achieve 97\% AUC value, an 81\% error reduction from its peers or similar RNNs without I-score.

\textbf{Future Research Directions.} We call for further exploration in the direction of using I-score to extrapolate features that have long-term dependencies in time-series and sequential data. Since it is computationally costly to rely on high-performance CPU/GPUs, the direction I-score is taking leads researchers to rethink about designing longer and deeper neural networks. Instead, future research can continue under the approach of building low dimensional but extremely informative features so that we can construct less complicated models in order for end-users.

\section*{Contribution}
\label{sec:contribution}

We would like to dedicate this to H. Chernoff, a well-known statistician and a mathematician worldwide, in honor of his 98th birthday and his contributions in Influence Score (I-score) and the Backward Dropping Algorithm (BDA). We are particularly fortunate in receiving many useful comments from him. Moreover, we are very grateful for his guidance on how I-score plays a fundamental role that measures the potential ability to use a small group of explanatory variables for classification which leads to much broader impact in fields of pattern recognition, computer vision, and representation learning.

\section*{Acknowledgement}
\label{sec:acknowledgement}

This research is supported by National Science Foundation BIGDATA IIS 1741191.

\bibliographystyle{unsrt}  
\bibliography{references}






\end{document}